\pdfoutput=1

\documentclass[11pt]{article}
\usepackage[final]{acl}

\usepackage{cmap}

\usepackage{times}
\usepackage{latexsym}
\usepackage[T1]{fontenc}
\usepackage[final]{microtype}
\usepackage{inconsolata}

\usepackage{fix-cm}
\usepackage{amsmath}
\usepackage{amssymb}
\usepackage{amsfonts}
\usepackage{graphicx}
\usepackage{relsize}
\usepackage{bbm}
\usepackage{booktabs}
\usepackage{multirow}
\usepackage{subcaption}

\allowdisplaybreaks

\title{
    ConGraT: Self-Supervised Contrastive Pretraining for \\
    Joint Graph and Text Embeddings
}

\author{
    William Brannon\thanks{\enspace Corresponding author.}\enspace\quad
    Wonjune Kang \AND
    Suyash Fulay \quad
    Hang Jiang \quad
    Brandon Roy\quad
    Deb Roy\quad
    Jad Kabbara
    \\\\
    MIT Center for Constructive Communication \\
    \texttt{\{wbrannon, wjkang, sfulay, hjian42, bcroy, dkroy, jkabbara\}@mit.edu}
}

\begin{document}

\maketitle

\begin{abstract}
Learning on text-attributed graphs (TAGs), in which nodes are associated with one or more texts, has been the subject of much recent work. However, most approaches tend to make strong assumptions about the downstream task of interest, are reliant on hand-labeled data, or fail to equally balance the importance of both text and graph representations. In this work, we propose \textbf{Con}trastive \textbf{Gra}ph-\textbf{T}ext pretraining (ConGraT), a general, self-supervised approach for jointly learning separate representations of texts and nodes in a TAG. Our method trains a language model (LM) and a graph neural network (GNN) to align their representations in a common latent space using a batch-wise contrastive learning objective inspired by CLIP. We further propose an extension to the CLIP objective that leverages graph structure to incorporate information about inter-node similarity. Extensive experiments demonstrate that ConGraT outperforms baselines on various downstream tasks, including node and text category classification, link prediction, and language modeling. Finally, we present an application of our method to community detection in social graphs, which enables finding more \textit{textually grounded} communities, rather than purely graph-based ones. Code and certain datasets are available at \url{https://github.com/wwbrannon/congrat}.
\end{abstract}
\section{Introduction}
\label{sec:intro}
Recent advances in multimodal representation learning have shown the benefits of simultaneously modeling language with other modalities, which allows for more efficient training and improved downstream performance of both sets of learned representations. These benefits have been especially clear in text/vision or text/audio applications, which often see large improvements in predictive performance or generative modeling ability \citep{radfordLearningTransferableVisual2021, liSupervisionExistsEverywhere2021, muSLIPSelfsupervisionMeets2022, elizalde2023clap}. In this work, we address another modality that frequently co-occurs with text: network- or graph-structured data.

We consider in particular the scenario of a text-attributed graph (TAG); that is, a graph over entities (i.e., nodes) associated with one or more texts. Such graphs occur frequently in the real world; examples include social media graphs of users and their posts, link graphs over web pages and their content, and citation networks of articles or authors and the texts of academic articles. In this setting, rather than modeling each source of data separately, graph information may be used to improve performance on language tasks and text information may be leveraged for graph tasks such as link prediction or node classification.

Prior work has approached the problem of combining these two modalities in several ways. Some approaches have used textual data to inform or supervise training of graph neural networks (GNNs) \citep{yangNetworkRepresentationLearning2015, zhangUserProfilePreserving2017, liuContentNodeSelfTranslation2018, zhangTextGraphTransformer2020}, but these methods do not produce graph-informed text representations. This is more parameter-efficient for graph-only tasks, but means that separate modeling is needed to solve text-based tasks while leveraging graph data. Other works have considered the converse case of employing a TAG structure to fine-tune pretrained language models (PLMs) \citep{cohanSPECTERDocumentlevelRepresentation2020a, yasunagaLinkBERTPretrainingLanguage2022, ostendorffNeighborhoodContrastiveLearning2022}. Although these approaches allow for the extraction of graph-informed text embeddings, they have the opposite limitation to the above of not learning node representations.
While there have been attempts to learn joint representations of nodes and texts, they all have certain limitations, such as requiring a supervised objective and labeled data \citep{liEncodingSocialInformation2019, chandraGraphbasedModelingOnline2020}, freezing either the text or graph embeddings/encoders \citep{gourruGaussianEmbeddingLinked2020, karpovSocialBERTTransformersOnline2022a}, or relying on the particular structure of one application \citep{liJointEmbeddingModels2017}.
Several recent works have leveraged joint training of PLMs with GNNs to integrate both text and graph information for representation learning in each modality \citep{yangGraphFormersGNNnestedTransformers2021, chienNodeFeatureExtraction2022, zhaoLearningLargescaleTextattributed2023}.
These methods, however, make specific modeling assumptions based on the tasks that they aim to solve, employ complex training procedures that alternately optimize the PLM and GNN modules, or need human-annotated knowledge distillation, which in general go against the goal of self-supervised learning.

\begin{figure}[t]
    \centering
    \includegraphics[width=0.5\textwidth]{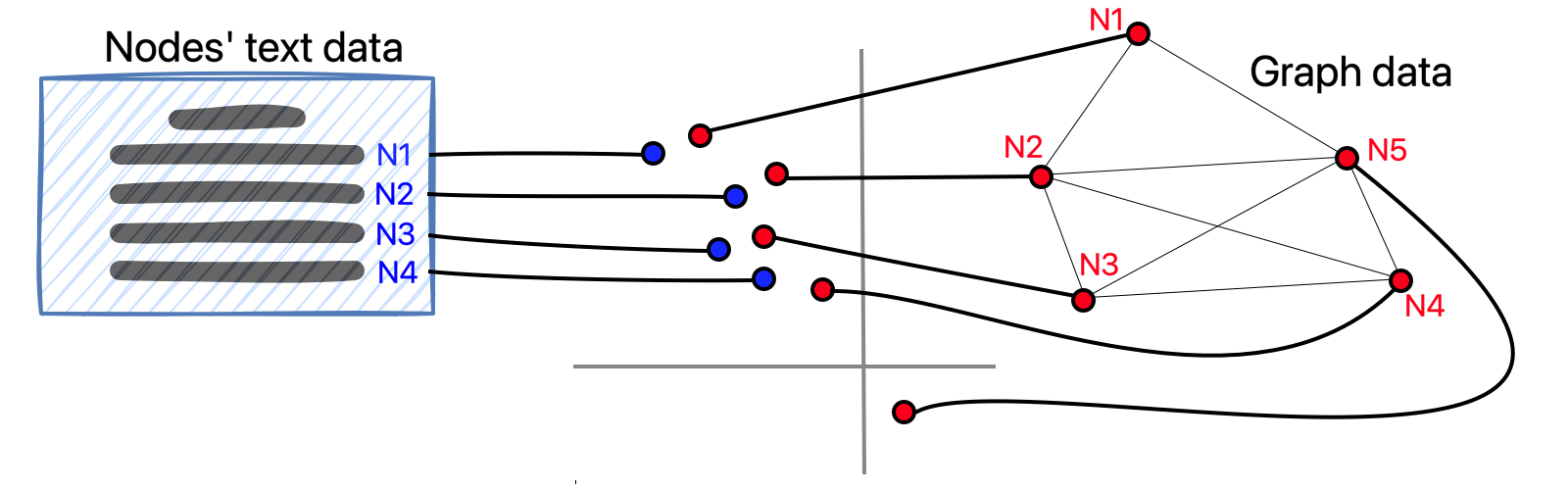}
    \caption[Illustration of joint embeddings]{Embeddings of graph nodes in red (e.g., Twitter users), and their associated texts in blue (e.g., tweets). They are placed into a common embedding space, with nodes near their associated texts. Node-text pairs are labeled N1 to N5. Note that not every node must have an associated text (here, N5 does not).}
    \label{fig:joint-embedding-illustration}
\end{figure}

In this work, we propose ConGraT (\textbf{Con}trastive \textbf{Gra}ph-\textbf{T}ext pretraining), a general approach to self-supervised joint graph-text learning based on a batch-wise contrastive learning objective inspired by CLIP \citep{radfordLearningTransferableVisual2021}. The idea is to have separate encoders for texts and graph nodes (more specifically, a PLM and a GNN, respectively) that are trained to align their representations within a common latent space, as shown in \autoref{fig:joint-embedding-illustration}. Taking advantage of the fact that graphs have greater structure than images, we propose an extension to the CLIP objective that incorporates information about plausible ``next guesses'' based on graph similarity. Our objective also admits an interpretation as a continuous relaxation of the contrastive CLIP objective over each node's two-hop neighborhood.

ConGraT provides flexibility in the choice of text and graph encoders and does not make assumptions on the structure of the TAG or any downstream task. As illustrated in our experiments, it is also inductive \citep{hamiltonInductiveRepresentationLearning2017}, with the encoders being able to generalize to previously unseen graphs as well as previously unseen texts. Experiments on various datasets show that ConGraT models consistently outperform strong baselines on various downstream tasks such as node and text category classification and link prediction. Additionally, we analyze how joint training affects language modeling performance, finding that ConGraT also results in improvements on this task on all datasets.

The contributions of this work are threefold: 1) We propose ConGraT, a general self-supervised pretraining method for jointly learning graph node and text representations on TAGs, such as citation, link, or social graphs. 2) We demonstrate that our joint pretraining method improves performance over strong unimodal and cross-modal baselines on various downstream tasks. 3) We release our code and datasets, including in particular a version of the Pubmed \citep{senCollectiveClassificationNetwork2008} graph learning dataset fully rebuilt from ground-truth Pubmed APIs, which includes the text of titles and abstracts as well as network data.\footnote{\url{https://github.com/wwbrannon/congrat}}
\section{Related Work}
\label{sec:related}

\subsection{Text-Augmented GNNs}
Text data can be incorporated into the learning of GNN representations in various ways. For example, \citet{yangNetworkRepresentationLearning2015} extend the DeepWalk algorithm to incorporate text features into node representations. \citet{liuContentNodeSelfTranslation2018} develop a seq2seq framework which learns node embeddings with inputs based on texts associated with the nodes. \citet{tuCANEContextAwareNetwork2017} use a selective attention mechanism to generate text-informed node embeddings for particular social contexts. \citet{zhangUserProfilePreserving2017} leverage kernel methods to construct node representations from user profile information in a way that incorporates network structure. Other methods include extracting graphs from entity co-occurrence in texts and modeling them \citep{zhangTextGraphTransformer2020, wallerQuantifyingSocialOrganization2021}. However, these approaches are limited in that, while they learn to represent nodes, they do not also learn graph-informed text representations.

\subsection{Graph-Augmented PLMs}
Another line of work uses information from graph structures to inform finetuning or further training of PLMs. SPECTER \citep{cohanSPECTERDocumentlevelRepresentation2020a} contrastively finetunes a language model by augmenting it with a measure of inter-node relatedness, with positive and negative examples for a triplet loss selected according to citation graph edges. LinkBERT \citep{yasunagaLinkBERTPretrainingLanguage2022} uses a graph structure to assemble training samples for a masked language model, pairing anchor texts with texts from contiguous, linked, or random documents, and uses an auxiliary document relation prediction (DRP) objective.
SciNCL \citep{ostendorffNeighborhoodContrastiveLearning2022} relaxes a discrete citation graph into a continuous domain with nearest-neighbor sampling.
SocialBERT \citep{karpovSocialBERTTransformersOnline2022a} and LMSOC \citep{kulkarniLMSOCApproachSocially2021} condition or augment the inputs to PLMs with frozen node representations that the model can attend over.
The models these methods learn produce text embeddings for documents, but do not also generate text-informed node representations.

\subsection{Joint Learning of PLMs and GNNs on TAGs}
More recently, representation learning on TAGs that jointly leverages graph and text information has been growing in popularity. Prefix tuning \citep{liPrefixTuningOptimizingContinuous2021} is a lightweight way of learning node-specific linguistic information and generates dense node representations in the process; however, it takes no advantage of the graph structure over the nodes. For fixed text and graph encoders, one can learn mappings from their separate embedding spaces to a common one, such as by canonical correlation analysis \citep{guptaScientificArticleRecommendation2017}.
Other methods jointly train text and graph encoders using an externally supervised objective \citep{chandraGraphbasedModelingOnline2020} or tailored for certain tasks \citep{liEncodingSocialInformation2019, gourruGaussianEmbeddingLinked2020}.
However, these methods all address specific settings that are not generalizable to more diverse tasks.

GraphFormers \citep{yangGraphFormersGNNnestedTransformers2021} jointly train a GNN with a PLM so as to learn text-informed node representations. However, they require a complex progressive learning strategy that iteratively utilizes manipulated and raw data.
GIANT \citep{chienNodeFeatureExtraction2022} predicts graph structure using PLMs to provide better initial embeddings for GNNs. However, the language model embeddings cannot be jointly optimized during the GNN training phase. GLEM \citep{zhaoLearningLargescaleTextattributed2023} uses a variational expectation-maximization (EM) framework that alternately updates a PLM and GNN separately using pseudo-labels predicted by the other module. While it enables improved scalability, the training procedure is complex and relies on the availability of task-specific target labels. In contrast, ConGraT is a general representation learning method for both graph nodes and texts, applicable to any inductive or transductive setting without such assumptions or complex training paradigms.
\section{Methodology}
\label{sec:methodology}

We consider a directed or undirected TAG, each node of which is associated with a set of one or more texts. The goal is to learn a shared latent space that allows us to place the embeddings of nodes and texts in semantically meaningful locations within that space. Formally, let $G = (V, E)$ be a graph, with $V$ the set of nodes and $E \subseteq V \times V$ the set of edges. Also, let $T^{(v)} = \{t_i^{(v)}\}_{i = 1}^{N_v}$, for $v \in V$, be the set of node $v$'s texts, with $N_v$ the number of texts corresponding to node $v$. We model $t_i^{(v)}$, the $i$-th text of node $v$, as a finite sequence of tokens over a vocabulary $W$, where $L_i^{(v)}$ is the length of $t_i^{(v)}$: $t_i^{(v)} = (S_0, S_1, S_2, ..., S_{L_v^{(i)}})$. The first and last tokens are always special start and end tokens.

Our training framework involves a \textit{text encoder}, a function $F_T : \cup_{i = 1}^\infty \otimes_i W \rightarrow \mathbb{R}^d$ from the set of all token sequences to a $d$-dimensional Euclidean embedding space. Similarly, we have a \textit{node encoder}, a function $F_G : V \rightarrow \mathbb{R}^d$ from nodes to an embedding space of the same dimension. (Note that while its domain is nodes, not edges, $F_G$ also depends on the edge set $E$.)
We aim to train the two encoders such that they learn a joint latent space between the text and graph node embeddings. This will allow us to use geometric properties of a common space to relate nodes and texts to each other for downstream inferential purposes.

\subsection{Approach}
The text and node encoders in ConGraT (a PLM and GNN, respectively) are connected at the output layers by a batch-wise contrastive training objective inspired by CLIP \citep{radfordLearningTransferableVisual2021}. The encoders are trained to align their representations in a joint latent embedding space. 
As in CLIP, each encoder is set behind an adapter module which generates embeddings of the same dimension.
Each adapter consists of two fully connected layers with a GeLU activation \citep{hendrycksGaussianErrorLinear2020} in between, followed by layer normalization \citep{baLayerNormalization2016}, and dropout \citep{srivastavaDropoutSimpleWay2014}.
This approach is flexible and allows use of many different kinds of both text and node encoders. On the text side, we illustrate this flexibility with experiments employing both causal and masked PLMs.

\begin{figure*}[!htb]
    \centering
    \includegraphics[width=0.76\textwidth]{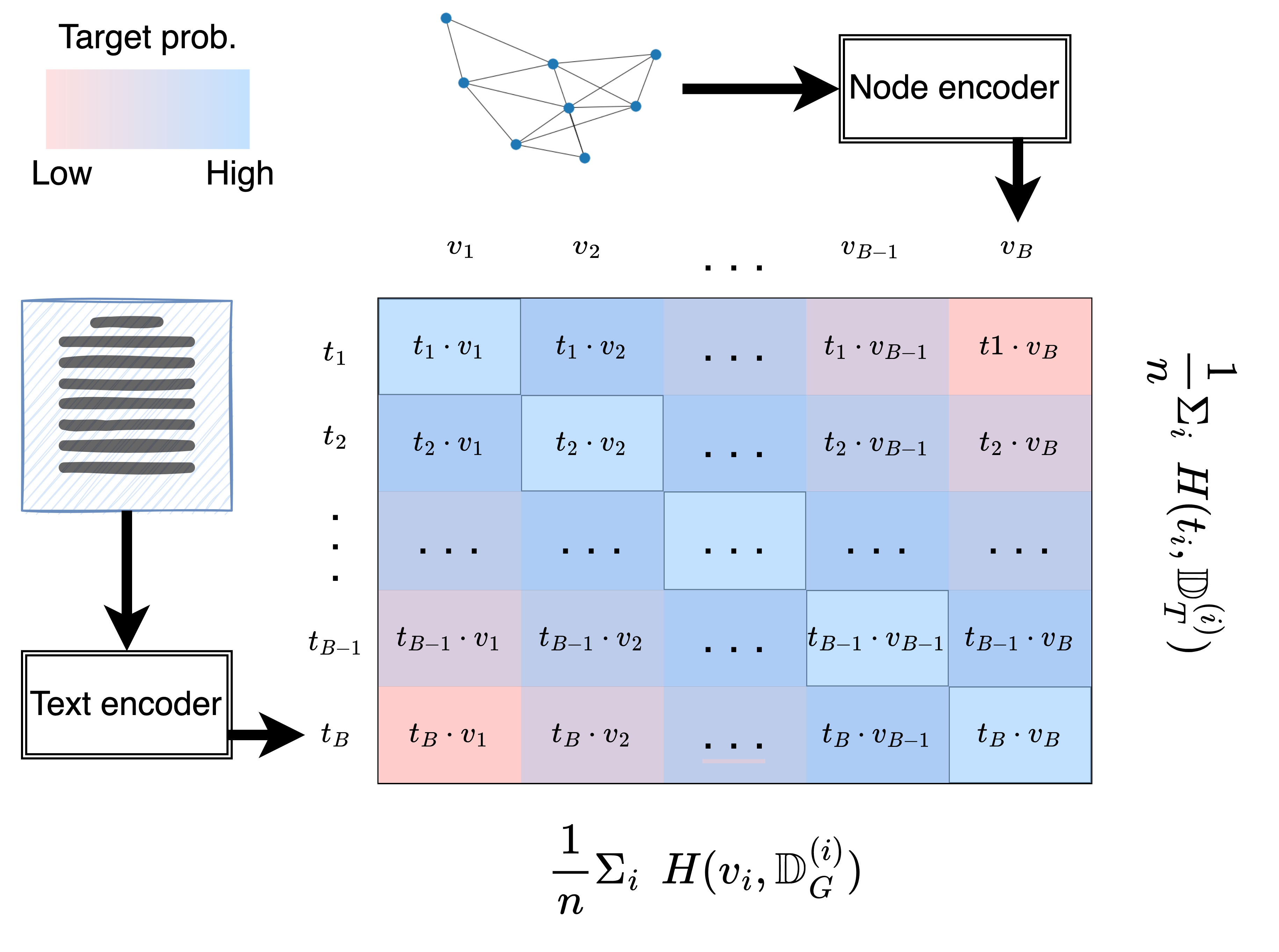}
    \vspace{-10pt}
    \caption{The overall architecture of our model. Given a minibatch of (text, origin node) pairs, node and text embeddings are generated by their respective encoders, then used to compute pairwise cosine similarities. The final loss is the average of cross entropies along each row and column of the similarity matrix, with each row $i$'s target probabilities (labeled $\mathbb{D}_T^{(i)}$ and $\mathbb{D}_G^{(i)}$) a mixture of the true targets (on the diagonal) and a (row- or column-specific) distribution proportional to a graph-based similarity measure.}
    \label{fig:architecture}
\end{figure*}

\paragraph{Training objective.}
We augment the standard InfoNCE loss \citep{oordRepresentationLearningContrastive2019} in CLIP with additional graph-specific elements. Unlike the vision-language case, in graphs, there are easily computable measures of how similar pairs of nodes are, such as their SimRank \citep{jehSimRankMeasureStructuralcontext2002} or their number of mutual in- and out-edges. We use these measures to incorporate information about the most likely second, third, and further choices for the nodes a text may originate from as well as the texts that may be associated with a node. The method is visualized in \autoref{fig:architecture}.

More formally, let $X = \cup_{v \in V} \{(v, t_i^{(v)})\}_{i = 1}^{|T^{(v)}|}$ be a dataset of (node, text) pairs, and let $B = \{(v_i, t_i^{(v_i)})\}_{i = 1}^{N_{B}}\subseteq X$ be a minibatch of size $N_B$ sampled from $X$. Now, fix an ordering of nodes, with $v_j$ the $j$-th node. Then, in terms of the text and node encoders $F_T$ and $F_G$, the matrix $C$ given by
$$
C_{ij} \triangleq e^\tau \frac{F_T(t_i^{(v_i)}) \cdot F_G(v_j)}{ \lVert F_T(t_i^{(v_i)}) \rVert \cdot \lVert F_G(v_j) \rVert}
$$
is the $N_B \times N_B$ matrix of cosine similarities between texts and nodes in the batch. (Note that $C$ is square but not symmetric: rows are texts and columns are nodes.) The matrix is multiplied by a scalar factor $e^\tau$, where $\tau$ is a log-temperature parameter that allowing some learnable control over the learning rate, reducing sensitivity to the choice of learning rate. We empirically initialize $\tau = 3.5$ based on our experiments (see \autoref{sec:appendix-models-training}).

Further, let $S_T(\cdot, \cdot)$ and $S_G(\cdot, \cdot)$ be graph-based \textit{similarity functions} for texts and nodes, respectively, assigning non-negative continuous similarity scores. Then, we define graph-based similarity distributions for texts and nodes. 
Where $K_T(i) = \sum_{k=1}^{N_B} S_T(t_i^{(v_i)}, t_k^{(v_k)})$ and $K_G(i) = \sum_{k=1}^{N_B} S_G(v_i, v_k)$, then $\forall i, j$, we have
$$
s_T^{(i)}(j) = \frac{S_T(t_i^{(v_i)}, t_j^{(v_j)})}{K_T(i)},
s_G^{(i)}(j) = \frac{S_G(v_i, v_j)}{K_G(i)}.
$$
The target distributions are mixtures of these distributions and indicator variables for the true source node of a text and matching text of a node. For each example $X_j = (v_j, t_j^{(v_j)})$ in the minibatch, fixing some hyperparameter $\alpha \in [0, 1]$, we define the target distribution in the text case by $\mathbb{D}_T^{(j)}(\alpha) = (1 - \alpha) \mathbbm{1}_i\{v_i = v_j\} + \alpha s_T(j)$, where $s_T(j)$ is the vector of $s^{(i)}_T(j)$ values for all $i$. In the graph case, where $s_G(j)$ is defined analogously, we have $\mathbb{D}_G^{(j)}(\alpha) = (1 - \alpha) \mathbbm{1}_i\{t_i = t_j\} + \alpha s_G(j)$.
Then where $H$ is the cross-entropy and $C_{i,:}$, $C_{:,i}$ are the $i$-th row and $i$-th column of $C$, our loss is
$$
\mathlarger{\mathcal{L}(B; \alpha) =\frac{1}{2N_B} \mathlarger{\mathlarger{\sum}}}_{i = 1}^{N_B}
\begin{aligned}
& \: H(C_{i,:}, \mathbb{D}_T^{(i)}(\alpha)) \; \\
& \: + H(C_{:,i}, \mathbb{D}_G^{(i)}(\alpha)).
\end{aligned}
$$
 
With $\alpha = 0$, this loss is equivalent to the average of cross-entropies for predictions of which node in the minibatch goes with which text and which text goes with which node. With higher values of $\alpha$, the target distributions are mixtures of indicators for the true source node and text and the distribution of other nodes and texts by graph similarity. If similar nodes produce similar texts, as suggested by the homophily principle \citep{dechoudhuryBirdsFeatherDoes2010}, positive $\alpha$ values should allow the model to learn more efficiently. Even if not all graph nodes are closely related to their texts, this objective should be able to learn from those that are.

\paragraph{Similarity function.}
For undirected graphs, we base our similarity function on a node pair's number of mutual neighbors. If $A$ is the graph adjacency matrix, we compute $AA^T$ to find the number of mutual in- or out-neighbors of each node pair, and find the cosine similarity of each row $i$ and column $j$ of $AA^T$ to measure the similarity of nodes $i$ and $j$. A benefit of this function over alternatives like SimRank \citep{jehSimRankMeasureStructuralcontext2002} is its lower computational cost and faster runtime for large graphs. On the text side, because we are interested in leveraging graph information, we approximate the similarity of a pair of texts as that of the associated nodes. The digraph case is more complicated, as it requires a directed similarity function that can distinguish between edges $(i, j)$ and $(j, i)$. We defer choosing and validating such a function to future work; thus, all experiments with $\alpha > 0$ in \autoref{sec:experiments} discard edge directions.

\subsection{Theoretical View}
From a theoretical perspective, the above similarity function allows our training objective to be viewed as a continuous relaxation of the contrastive CLIP objective across a node's two-hop neighborhood. (Only nodes with shared neighbors, which are in each other's two-hop neighborhoods, will have positive similarity.)  
Different choices of similarity correspond to different choices of how to relax the contrastive objective across the graph; in particular, restricting to the one-hop neighborhood amounts to using a coarse, binary indicator of similarity, with $S_G(u, v)$ the indicator function for edge $(u, v)$.

This view indicates a connection to label smoothing \citep{szegedy2015rethinking}, but with the smoothing distribution based on graph similarity rather than being uniformly random. It also casts our model as an extension to a graph setting of prior work on similarity-based smoothing in non-graph contexts \citep{liu2021classsimilarity}. Note that this is a different sense of ``smoothing'' than the aggregation over neighbor representations that the term refers to in the context of message-passing GNNs.
\section{Experimental Setup}
\label{sec:experiments}
We evaluate our approach on three tasks: node category classification, link prediction, and language modeling. Link prediction and language modeling are fundamental modality-specific metrics that measure how well our node and text encoders retain the ability to model their individual modalities. We perform node classification using each encoder's embeddings in order to measure how effective the learned representations are for downstream tasks. Further details are provided in \autoref{sec:appendix-models-training}.

\subsection{Datasets}
\label{subsec:experiments-setup}

We evaluate on three datasets, comprising one each of citation, link, and social graphs: 
(1) the Pubmed dataset~\citep{senCollectiveClassificationNetwork2008}, (2) a dataset of Wikipedia articles represented by their introductory paragraphs and the hyperlinks between the articles in those paragraphs, selected from the broader T-REx Corpus \citep{elsaharTRExLargeScale2018}, and (3) a novel Twitter dataset comprising high-profile public figures, which includes the tweets, the follow graph over the associated users, and some demographic information about them (age, gender, United States political party, etc.).
We include the last Twitter dataset to demonstrate the performance of our method on social network TAGs, which is a setting that has been less explored in prior work on joint graph and language learning.
\autoref{tab:dataset-statistics} shows descriptive statistics of the datasets. Because we use entirely separate train, validation, and test splits, without shared graph structure, our results below are in an inductive (rather than transductive) setting.
More information about the datasets and our collection procedures for Twitter data and the other datasets' raw text are provided in \autoref{sec:appendix-datasets}.

\subsection{Models}
For each dataset, we train two ConGraT variants with masked and causal PLMs, initializing with weights from MPNet \citep{songMPNetMaskedPermuted2020} and DistilGPT-2 \citep{sanhDistilBERTDistilledVersion2019}, respectively. Specifically, we use the pretrained \texttt{all-mpnet-base-v2} and \texttt{distilgpt2} models from the \texttt{sentence-transformers} toolkit \citep{ reimersSentenceBERTSentenceEmbeddings2019}.
For the graph node encoder, we use a graph attention network (GAT) \citep{velickovicGraphAttentionNetworks2018} with 3 layers of 2 attention heads each, randomly initialized and trained from scratch. All text and graph embeddings have dimension 768, and we obtain text-level representations from the PLM text encoder by mean pooling.

We examine models with ($\alpha = 0.1$) and without ($\alpha = 0.0$) graph similarity information included in the loss. We also examine models which consider edge directions (and thus have $\alpha = 0.0$).\footnote{Recall that because we defined a similarity function with $\alpha > 0$ only for undirected graphs, there are no experiments with directed edges where $\alpha = 0.1$.} In all, between these three factors (masked or causal PLM, $\alpha = 0.0$ or $\alpha = 0.1$, directed or undirected edges), and without experiments with $\alpha = 0.1$ for directed edges, there are 6 possible model combinations on each dataset, for a total of 18 combinations.

\begin{table}[t]
\centering
\setlength{\tabcolsep}{6pt}
\begin{tabular}{lcccc}
    \toprule
    & Pubmed    & T-REx     & Twitter \\
    \midrule
    \# Nodes    & 19,716    & 9,214     & 8,721 \\
    \# Edges    & 61,110    & 22,689    & 2,373,956 \\
    \# Texts    & 59,381    & 18,422    & 167,558 \\
    \# Classes  & 3         & 5         & 13 (5 tasks) \\
    \bottomrule
\end{tabular}

\caption{Statistics for the Pubmed, T-REx, and Twitter datasets used in our experiments.}
\label{tab:dataset-statistics}
\end{table}

\subsection{Baselines}

For node representations, we compare against embeddings from a GNN-only baseline: a unimodal GAT autoencoder with the same architecture as the ConGraT node encoder, trained as usual to reconstruct the adjacency matrix without added similarity information. For text representations, in addition to unimodal masked and causal PLM baselines finetuned on each dataset, we also compare against two models leveraging both modalities: LinkBERT \citep{yasunagaLinkBERTPretrainingLanguage2022} and Social-LM, a modified implementation of SocialBERT \citep{karpovSocialBERTTransformersOnline2022a} and LMSOC \citep{kulkarniLMSOCApproachSocially2021}. Because LinkBERT uses a masked language modeling objective, it is used as a baseline only for the masked versions of ConGraT. Initial node representations for all GNN models are sentence embeddings of text associated with each node: for Pubmed, the concatenated text of the title and abstract sections; for Twitter, user bios; for T-REx, the Wikipedia article text. Further implementation details are given in \autoref{sec:appendix-models-training}.

\begin{table}[t]
\centering
\setlength{\tabcolsep}{4pt}
\footnotesize\begin{tabular}{llcccc}
\toprule
 &  & \multicolumn{2}{c}{Pubmed} & \multicolumn{2}{c}{T-REx} \\
 \cmidrule(lr){3-4} \cmidrule(lr){5-6}
 &  & C & M & C & M \\
\midrule
\multirow[c]{3}{*}{\rotatebox[origin=c]{90}{Graph}} & ConGraT ($\alpha = 0$) & 0.967 & \bfseries 0.964 & \bfseries 0.951 & 0.937 \\
 & ConGraT ($\alpha = 0.1$) & \bfseries 0.973 & 0.963 & 0.949 & \bfseries 0.946 \\
 \cmidrule(lr){2-6}
 & GAT & 0.956 & 0.956 & 0.939 & 0.939 \\
\midrule
\multirow[c]{5}{*}{\rotatebox[origin=c]{90}{Text}} & ConGraT ($\alpha = 0$) & 0.962 & 0.958 & 0.920 & 0.911 \\
 & ConGraT ($\alpha = 0.1$) & \bfseries 0.969 & \bfseries 0.966 & \bfseries 0.931 & \bfseries 0.928 \\
 \cmidrule(lr){2-6}
 & LinkBERT & -- & 0.954 & -- & 0.906 \\
 & Social-LM & 0.858 & 0.878 & 0.890 & 0.851 \\
 & Unimodal LM & 0.931 & 0.943 & 0.908 & 0.892 \\
\bottomrule
\end{tabular}
\caption{Node classification performance (test-set AUC) on article labels of the Pubmed and T-REx datasets. For T-REx, we show the average AUC over the category labels because the dataset is multilabel rather than multiclass. \textbf{Bold} values mark best-performing models. C = causal, M = masked.}
\label{tab:classification-txt-pubmed-trex}
\end{table}

\begin{table*}[!htb]
\centering
\setlength{\tabcolsep}{7pt}
\small{
    \begin{tabular}{llcccccccccccc}
    \toprule
     &  & \multicolumn{2}{c}{Age} & \multicolumn{2}{c}{Gender} & \multicolumn{2}{c}{Occupation} & \multicolumn{2}{c}{Party} & \multicolumn{2}{c}{Region} \\
     \cmidrule(lr){3-4} \cmidrule(lr){5-6} \cmidrule(lr){7-8} \cmidrule(lr){9-10} \cmidrule(lr){11-12}
     &  & C & M & C & M & C & M & C & M & C & M \\
    \midrule
    \multirow[c]{3}{*}{\rotatebox[origin=c]{90}{Graph}} & ConGraT ($\alpha = 0$) & 0.646 & 0.665 & \bfseries 0.811 & \bfseries 0.802 & \bfseries 0.993 & 0.989 & \bfseries 0.966 & 0.959 & \bfseries 0.755 & \bfseries 0.780 \\
     & ConGraT ($\alpha = 0.1$) & \bfseries 0.650 & \bfseries 0.682 & 0.803 & 0.801 & 0.992 & \bfseries 0.993 & 0.960 & \bfseries 0.986 & 0.742 & 0.774 \\
     \cmidrule(lr){2-12}
     & GAT & 0.631 & 0.631 & 0.713 & 0.713 & 0.967 & 0.967 & 0.757 & 0.757 & 0.678 & 0.678 \\
    \midrule
    \multirow[c]{5}{*}{\rotatebox[origin=c]{90}{Text}} & ConGraT ($\alpha = 0$) & \bfseries 0.622 & \bfseries 0.628 & 0.663 & \bfseries 0.668 & \bfseries 0.961 & \bfseries 0.959 & \bfseries 0.771 & 0.787 & \bfseries 0.693 & 0.679 \\
     & ConGraT ($\alpha = 0.1$) & 0.620 & 0.624 & \bfseries 0.668 & 0.661 & 0.960 & 0.958 & 0.771 & \bfseries 0.796 & 0.686 & \bfseries 0.680 \\
     \cmidrule(lr){2-12}
     & LinkBERT & -- & 0.617 & -- & 0.661 & -- & 0.954 & -- & 0.762 & -- & 0.606 \\
     & Social-LM & 0.566 & 0.567 & 0.602 & 0.608 & 0.921 & 0.909 & 0.628 & 0.676 & 0.582 & 0.572 \\
     & Unimodal LM & 0.610 & 0.613 & 0.649 & 0.655 & 0.948 & 0.945 & 0.742 & 0.769 & 0.587 & 0.598 \\
    \bottomrule
    \end{tabular}
}
\caption{Node classification performance (test-set AUC) on user traits from the Twitter dataset. \textbf{Bold} values mark best-performing models. C = causal, M = masked.
}
\label{tab:classification-txt-twitter}
\end{table*}
\section{Results}
\label{subsec:experiments-results}

\subsection{Node Classification}
We train logistic regression models to perform node classification on the Pubmed, Twitter, and T-REx datasets using the embeddings generated from each ConGraT model or baseline. Overall, ConGraT models achieve high performance on this set of tasks relative to baselines on all three datasets.

\autoref{tab:classification-txt-pubmed-trex} shows AUCs for article label classification on Pubmed and T-REx, and \autoref{tab:classification-txt-twitter} shows AUCs for demographic classification tasks on age, gender, occupation, party, and region on users in our Twitter dataset. (At the text level, the dependent variable is that of the corresponding node.) The best ConGraT model achieves the highest node classification performance on all datasets, and the differences from the nearest baseline are statistically significant by a bootstrap test ($p < 10^{-4}$) in all cases. Even the less performant ConGraT model outperforms all baselines in 26 of 28 experiments.

\begin{figure}[t]
    \centering
    \begin{subfigure}[b]{0.22\textwidth}
        \centering
        \includegraphics[width=\textwidth]{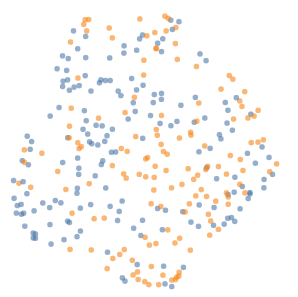}
        \caption{GAT}
        \label{fig:embed-viz-gat}
    \end{subfigure}
    \hfill 
    \begin{subfigure}[b]{0.22\textwidth}
        \centering
        \includegraphics[width=\textwidth]{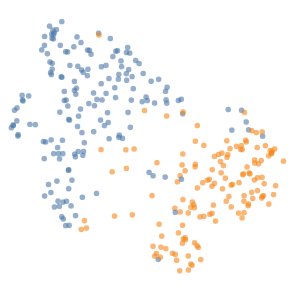}
        \caption{ConGraT}
        \label{fig:embed-viz-congrat}
    \end{subfigure}
    \caption{2D UMAP visualizations of GAT and ConGraT ($\alpha = 0.0$) embeddings on the Twitter data subset with U.S. political party labels (blue = Democrat, orange = Republican).}
    \label{fig:embed-viz}
\end{figure}

Notably, we see some of ConGraT's largest improvements when one modality has less signal than the other. For example, tweet text is less useful than graph data in predicting users' geographic region. Many Twitter edges are geographically nearby \citep{takhteyevGeographyTwitterNetworks2012}, and our method is more effective than baselines at infusing this information into an encoder which at inference time sees only text.

The more discriminative nature of representations learned by ConGraT can also be seen visually; \autoref{fig:embed-viz} shows a 2D UMAP plot comparing ConGraT and GAT embeddings on the Twitter data subset with U.S. political party labels, which validates that ConGraT embeddings have a much more clearly separated class boundary.

\subsection{Link Prediction}
We evaluate link prediction performance using inner product decoding \citep{kipfVariationalGraphAutoEncoders2016}
to derive edge existence probabilities from embeddings. As a baseline, we use the same GAT architecture as in our jointly trained ConGraT models and train it directly on link prediction using the same inner product decoding.

Results are shown in \autoref{tab:link-prediction}. All ConGraT models outperform the baselines, despite those baselines being specifically trained for link prediction. In some cases, these improvements are quite large, with the best-performing model on the Twitter dataset recording an AUC of 0.806 vs. 0.723 for the best-performing baseline, a relative increase of 11.5\%. Training with graph-based similarity information ($\alpha = 0.1$) often also leads to further improvements. Performance is similar for directed and undirected models, demonstrating our approach's adaptability to different types of graphs.
Notably, our model's high performance is zero-shot, with no additional training on link prediction.

\begin{table}[t]
\centering
\setlength{\tabcolsep}{5pt}
    \begin{tabular}{lcccccc}
    \toprule
     &  &  & Pubmed & T-REx & Twitter \\
    \midrule
    \multirow[c]{3}{*}{\rotatebox[origin=c]{90}{Masked}} & \multirow[c]{2}{*}{$\alpha = 0$} & U & 0.953 & 0.899 & 0.791 \\
     &  & D & 0.952 & 0.902 & 0.797 \\
    \cmidrule(lr){4-6}
     & $\alpha = 0.1$ & U & \bfseries 0.980 & \bfseries 0.951 & \bfseries 0.802 \\
    \midrule
    \multirow[c]{3}{*}{\rotatebox[origin=c]{90}{Causal}} & \multirow[c]{2}{*}{$\alpha = 0$} & U & 0.956 & 0.908 & \bfseries 0.806 \\
     &  & D & 0.950 & 0.897 & 0.799 \\
    \cmidrule(lr){4-6}
     & $\alpha = 0.1$ & U & \bfseries 0.979 & \bfseries 0.957 & 0.799 \\
    \midrule
    \multirow[c]{2}{*}{\rotatebox[origin=c]{90}{GAT}} &
    \multirow[c]{2}{*}{--} & U & 0.943 & 0.927 & 0.713 \\
     &  & D & 0.940 & 0.925 & 0.723 \\
    \bottomrule
    \end{tabular}
\caption{Link prediction performance (test-set AUC) by dataset. \textbf{Bold} values mark best-performing models. U = undirected edges, D = directed.
}
\label{tab:link-prediction}
\end{table}

\subsection{Language Modeling}
Previous works that jointly trained LMs with GNNs on TAGs \citep{yangGraphFormersGNNnestedTransformers2021, chienNodeFeatureExtraction2022, zhaoLearningLargescaleTextattributed2023} evaluated on node classification tasks using the representations learned by each module, but did not study in depth how joint training affected the LM component's capabilities. We perform this analysis, evaluating joint pretraining's impact on downstream language-modeling performance. To do this, we attach a randomly initialized LM head to the ConGraT text encoder and further train both the encoder and head on causal language modeling. We evaluate with perplexity, and thus limit evaluation to causal-LM variants of our model (those initialized from DistilGPT-2). As a baseline, we finetune the baseline DistilGPT-2 LM on each dataset's texts.

\autoref{tab:lm-perplexity-text} shows the mean perplexity of causal LMs trained using ConGraT with $\alpha = 0$ and $\alpha = 0.1$ compared against the causal LM baseline. For all datasets, LMs trained using ConGraT achieve consistently lower average perplexity, and these differences are statistically significant by a paired $t$-test at the $5\sigma$ level ($p < 5.7 \times 10^{-7}$).

\begin{table}[t]
    \centering
        \begin{tabular}{lccc}
            \toprule
            & Pubmed & T-REx & Twitter \\
            \midrule
            $\alpha = 0$ & 6.95 & \bfseries 15.99 & 16.08 \\
            $\alpha = 0.1$ & \bfseries 6.94 & 16.07 & \bfseries 16.07 \\ \midrule
            LM & 6.98 & 16.84 & 16.44 \\
            \bottomrule
        \end{tabular}
    \caption{Language modeling performance (mean perplexity) of the causal ConGraT models vs. a unimodal LM baseline. \textbf{Bold} marks each dataset's best model.
    }
    \label{tab:lm-perplexity-text}
\end{table}

\subsection{Application: Community Detection}
To illustrate ConGraT's broad usability in applications, we compare it to other methods of detecting communities in the Twitter data. As baselines, we use Louvain community detection \citep{blondelFastUnfoldingCommunities2008} on the follow graph, and a clustering-based approach on the GAT baseline's embeddings using UMAP \citep{mcinnes2020umap} and HDBSCAN \citep{McInnes2017}. For ConGraT, we use the same clustering approach with embeddings from the $\alpha = 0.0$ variant. We expect these methods will find different kinds of communities: while Louvain communities are entirely determined by graph structure, and the GAT baseline can take some advantage of text via use of sentence embeddings as initial node vectors, we expect ConGraT to be most able to infuse textual information into network communities.

Because we want to determine how informed each set of communities is by the text associated with the graph, we evaluate by predicting community labels from text embeddings. For each of the above community detection methods, we first compute the centroid of each node's text embeddings and label it with the user's community. Then, we fit a logistic regression model on the training split and predict the test set community label from these centroid text features.

\begin{figure}[t]
    \centering
    \includegraphics[width=0.45\textwidth]{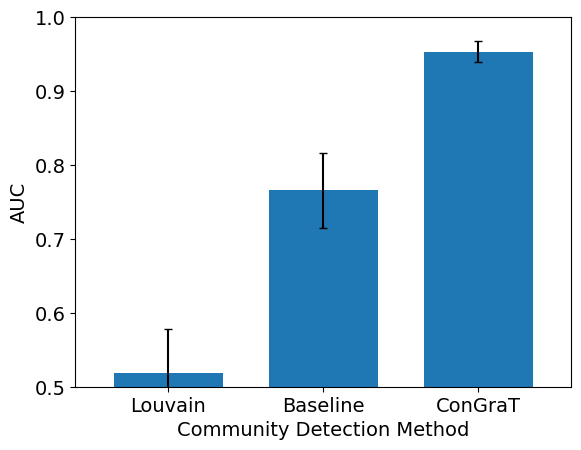}
    \caption[Community Detection]{Test-set AUCs for predictions of community labels from text embeddings on the Twitter dataset. ``Louvain'' denotes Louvain communities detected in the follow graph, ``Baseline'' the GAT baseline model, and ``ConGraT'' our model with $\alpha = 0.0$.}
    \label{fig:community-detection}
\end{figure}

The results, in \autoref{fig:community-detection}, demonstrate exactly the expected pattern: graph-based Louvain communities are poorly predictable from text, while communities clustered from baseline GAT embeddings are more predictable. The closest relationship to textual content occurs for communities detected with ConGraT embeddings. This pattern highlights a potential application of our method: detecting more \textit{discursively} or \textit{textually grounded} communities in social graphs, rather than ones based only on graph information (e.g., communities informed by discussion among political figures on Twitter).

\section{Conclusion}
\label{sec:conclusion}
We propose ConGraT (\textbf{Con}trastive \textbf{Gra}ph-\textbf{T}ext pretraining), a self-supervised pretraining framework for jointly learning text and graph node representations using pretrained language models (PLMs) and graph neural networks (GNNs) on a text-attributed graph (TAG). ConGraT uses a batch-wise contrastive learning objective to train text and graph encoders to align their representations within a common latent space. The framework is inductive, generalizable to any text or graph encoder architecture, and does not depend on the structure of the TAG or a particular downstream task. In experiments on citation, link, and social graphs, our method outperforms baselines on various downstream tasks, including node classification, link prediction, and language modeling. Our results also highlight the value of incorporating graph structure into our contrastive learning objective, with nonzero values of the $\alpha$ parameter often improving performance. Finally, an application to community detection, in which our method finds more textually grounded communities than alternative methods, highlights the broad applicability of this form of representation learning to many domains.

\section{Ethical Considerations}
\label{sec:ethics}
Applying this method to existing social or other networks poses the risk of learning and reproducing biases or toxic behavior exhibited in those datasets \cite{liangUnderstandingMitigatingSocial2021}. The risk of learning such harmful behavior is likely to be greatest on social media datasets, given the greater prevalence of harassment and toxic or ``antisocial'' behavior there \cite{saveskiStructureToxicConversations2021, atskeStateOnlineHarassment2021}. On the other hand, for applications like detecting hate speech and toxicity, this may be the intended behavior; careful attention to the ethics of how any model is used is key.

On the dataset side, we believe there are no meaningful concerns about release of personally identifying information (PII) or offensive content with the Pubmed or T-REx datasets, which are already public and have been vetted by the academic community. For the Twitter dataset, because the tweets themselves are public, the users included are public figures, and our demographic data are collected from public Wikipedia and Ballotpedia data, we do not believe that our use of the data poses privacy concerns either.
\section{Limitations}
\label{sec:limitations}
In this work, we make the assumption that there is a relationship between the graph structure over nodes and the texts associated with each node. If the two modalities are generated independently, or otherwise do not convey any useful information about each other, joint pretraining should not be expected to improve performance.

A more practical limitation is around the scale of graph data. Current graph neural networks require large amounts of memory, and scaling to massive graphs \cite{serafiniScalableGraphNeural2021a, maGraphNeuralNetworks2022} is still an active research area. Our approach thus requires modifications to scale to extremely large graphs. Sampling-based approaches like neighbor sampling \cite{hamiltonInductiveRepresentationLearning2017} are particularly promising and can reduce the memory of the graph encoder on each minibatch.

A further limitation of the present paper's analysis is that we have not validated a directed analogue of the similarity function for use with positive values of $\alpha$ in directed graphs. While one has the option of either using $\alpha = 0$ or discarding edge directions, we leave this issue for future work.
\subsubsection*{Acknowledgements}
\label{sec:acknowledgements}
The authors gratefully acknowledge support in the form of access to data from Twitter, Inc. We are also indebted to colleagues at the MIT Center for Constructive Communication for their feedback on earlier versions of this work.

\clearpage
\bibliography{references}

\clearpage
\appendix
\section{Sensitivity Analysis on \texorpdfstring{$\alpha$}{alpha}}
\label{sec:sensitivity}
To examine the hyperparameter $\alpha$'s impact on downstream performance, we conduct a sensitivity analysis on all three evaluation tasks, using the Pubmed dataset, with $\alpha = 0, 0.1, 0.5,$ and $1.0$. We use only the causal model variant for the LM task.

\begin{table}[ht]
\centering
\setlength{\tabcolsep}{5pt}
    \begin{tabular}{ccccc}
        \toprule
                       & NCG & NCT & LP & LM \\
        \midrule
        $\alpha$ = 0.0 & 0.967        & 0.962       & 0.956        & 6.95 \\
        $\alpha$ = 0.1 & \textbf{0.973}        & \textbf{0.969}       & \textbf{0.979}        & 6.94 \\
        $\alpha$ = 0.5 & 0.962        & 0.958       & 0.977        & 6.98 \\
        $\alpha$ = 1.0 & 0.941        & 0.900       & 0.897        & \textbf{6.88} \\
        \midrule
        Baseline       & 0.956        & 0.931       & 0.943        & 6.98 \\
        \bottomrule
    \end{tabular}
\caption[Sensitivity analysis]{Results of sensitivity analysis. NCG = node classification, graph; NCT = node classification, text; LP = link prediction; LM = language modeling. Values are AUC for the first three columns and perplexity for language modeling.}
\label{tab:sensitivity}
\end{table}

We find that $\alpha$'s impact varies by task. For link prediction and node classification, we see an intuitive pattern: Performance is best for $\alpha$ between 0 and 1, especially compared to $\alpha = 1$. That is, both components of our objective---matching nodes to texts, and matching nodes and texts to similar nodes and texts---add value. This pattern is not universal, however; while $\alpha > 0$, particularly $\alpha = 0.1$, consistently outperforms $\alpha = 0$, LM performance is best with $\alpha = 1.0$. We conjecture this pattern may be due to inter-document similarity in language use, which $\alpha = 1.0$ more effectively trains into the PLM. Overall, our results suggest both that $\alpha = 0.1$ is a reasonable default and that it may be worth tuning this parameter in practice.
\section{Embedding Space Geometry Analysis}
\label{sec:appendix-embedding-geometry}
To complement evaluation on downstream applications like node classification, link prediction and language modeling, this section pursues certain analyses of the geometry of the joint embedding space. We compare these jointly learned embedding spaces to the null model of separate spaces, as learned by the unimodal LM and GAT baselines.

We expect in particular that ConGraT's joint pretraining should align the two embedding spaces with each other, as well as with non-embedded distance metrics based entirely on the graph. We focus on examining how these distances (in the embedding spaces and the graph) relate to each other, because the geometric properties of a metric space are chiefly determined by the underlying metric. A finding of significantly increased alignment between distance metrics would indicate the models are effectively integrating information across language and graph modalities. As discussed below and shown in \autoref{tab:summary-eval-txt}, this is in fact exactly what we do see.

\paragraph{Inter-Embedding Distance Correlation.} First, we examine the correlation of the distance between pairs of texts with the distance between the corresponding pairs of nodes. That is, we sample text pairs $(t^{(u)}_1, t^{(v)}_2)$ from nodes $u$ and $v$, and examine over many such samples the correlation between the text-embedding distance $d_T(t^{(u)}_1, t^{(v)}_2)$ and the graph-embedding distance $d_G(u, v)$. We operationalize both in practice as the cosine distance.

With ConGraT pretraining, the cosine distance between text embeddings is substantially more correlated than in the separately trained case with the distance between the parent nodes. We see this effect for all model variants against the separate-spaces baseline on all datasets, and the increases are significant by a bootstrap test ($p < 10^{-6}$). Supporting our hypothesis, the two spaces have become systematically more aligned geometrically.

\paragraph{Embedding-Graph Distance Correlation.} Next, we relate the cosine distance in the text embedding space to a purely graph-based distance --- SimRank \cite{jehSimRankMeasureStructuralcontext2002} in our case. This extends the previous analysis by grounding the text-embedding distance more directly in the graph. In 34 out of 36 cases, we observe a significant increase over the separate-spaces baseline in the correlation between the text embedding distance and graph SimRank (at the $p = 10^{-6}$ level, using a similar bootstrap test).

\begin{table*}[!htb]
\centering
\begin{tabular}{lllccccc}
    \toprule
    \multirow[c]{2}{*}{Dataset} &
    \multirow[c]{2}{*}{Directed} & 
    \multirow[c]{2}{*}{LM Type} & 
    \multirow[c]{2}{*}{Sim.} &
    \multicolumn{2}{c}{Inter-Embedding} &
    \multicolumn{2}{c}{Text Emb.-Graph} \\
    \cmidrule(lr){5-6} \cmidrule(lr){7-8}
     &  &  &  & Joint & Separate & Joint & Separate \\
    \midrule
    \multirow[c]{6}{*}{Pubmed} & \multirow[c]{2}{*}{Directed} & Causal & $\alpha = 0.0$ & \bfseries 0.682 & 0.100 & \bfseries 0.118 & 0.019 \\
     &  & Masked & $\alpha = 0.0$ & \bfseries 0.604 & 0.248 & \bfseries 0.120 & 0.059 \\
     \cmidrule{2-8}
     & \multirow[c]{4}{*}{Undirected} & \multirow[c]{2}{*}{Causal} & $\alpha = 0.0$ & \bfseries 0.670 & 0.109 & \bfseries 0.157 & 0.026 \\
     &  &  & $\alpha = 0.1$ & \bfseries 0.679 & 0.109 & \bfseries 0.171 & 0.026 \\
     &  & \multirow[c]{2}{*}{Masked} & $\alpha = 0.0$ & \bfseries 0.603 & 0.260 & \bfseries 0.155 & 0.080 \\
     &  &  & $\alpha = 0.1$ & \bfseries 0.647 & 0.260 & \bfseries 0.173 & 0.080 \\
    \midrule
    \multirow[c]{6}{*}{TRex} & \multirow[c]{2}{*}{Directed} & Causal & $\alpha = 0.0$ & \bfseries 0.650 & 0.038 & \bfseries 0.131 & 0.022 \\
     &  & Masked & $\alpha = 0.0$ & \bfseries 0.564 & 0.248 & \bfseries 0.179 & 0.078 \\
     \cmidrule{2-8}
     & \multirow[c]{4}{*}{Undirected} & \multirow[c]{2}{*}{Causal} & $\alpha = 0.0$ & \bfseries 0.647 & 0.040 & \bfseries 0.215 & 0.027 \\
     &  &  & $\alpha = 0.1$ & \bfseries 0.704 & 0.040 & \bfseries 0.302 & 0.027 \\
     &  & \multirow[c]{2}{*}{Masked} & $\alpha = 0.0$ & \bfseries 0.600 & 0.248 & \bfseries 0.220 & 0.142 \\
     &  &  & $\alpha = 0.1$ & \bfseries 0.666 & 0.248 & \bfseries 0.272 & 0.142 \\
    \midrule
    \multirow[c]{6}{*}{Twitter} & \multirow[c]{2}{*}{Directed} & Causal & $\alpha = 0.0$ & \bfseries 0.319 & 0.035 & \bfseries 0.048 & 0.019 \\
     &  & Masked & $\alpha = 0.0$ & \bfseries 0.270 & 0.084 & \bfseries 0.049\textdagger & 0.047 \\
     \cmidrule{2-8}
     & \multirow[c]{4}{*}{Undirected} & \multirow[c]{2}{*}{Causal} & $\alpha = 0.0$ & \bfseries 0.317 & 0.036 & \bfseries 0.041 & 0.018 \\
     &  &  & $\alpha = 0.1$ & \bfseries 0.301 & 0.036 & \bfseries 0.048 & 0.018 \\
     &  & \multirow[c]{2}{*}{Masked} & $\alpha = 0.0$ & \bfseries 0.300 & 0.083 & 0.037 & \bfseries 0.044\textdagger \\
     &  &  & $\alpha = 0.1$ & \bfseries 0.226 & 0.083 & \bfseries 0.052 & 0.044 \\
    \bottomrule
    \end{tabular}
\caption{Correlations between pairs of distances as discussed in \autoref{sec:appendix-embedding-geometry}: those of the text and graph embedding spaces (``Inter-Embedding''), on the one hand, and the text embedding space and the graph-based SimRank distance (``Text Emb.-Graph''), on the other. The ``Joint`` column indicates the jointly trained embedding spaces from our ConGraT models, and the ``Separate'' column indicates the separately trained embedding spaces of the GAT and LM baselines. The most closely aligned pair of distances in each comparison, joint or baseline, is shown in \textbf{bold}. Differences marked with a \textdagger\, are not significant at the $p = 10^{-6}$ level by a bootstrap test.}
\label{tab:summary-eval-txt}
\end{table*}

\subsection{Retrieval}
\label{subsec:appendix-embedding-geometry-retrieval}
Finally, as an additional test of geometric alignment and cross-modal data integration, we consider a simple retrieval task: identifying the node associated with a given text. For each text, we select the node whose embedding has the highest cosine similarity to the text's embedding, and report the top-$k$ accuracy for $k$ from 1 to 10. This task might itself be an important downstream measure in a retrieval setting, but for purposes of geometric analysis we consider only the comparison to separate embedding spaces here. Note that as with CLIP, this use of our representations can be thought of as zero shot transfer for text or node classification (where objects in the other modality are the classes).

\begin{figure*}[htb]
    \centering
    \includegraphics[scale=0.51]{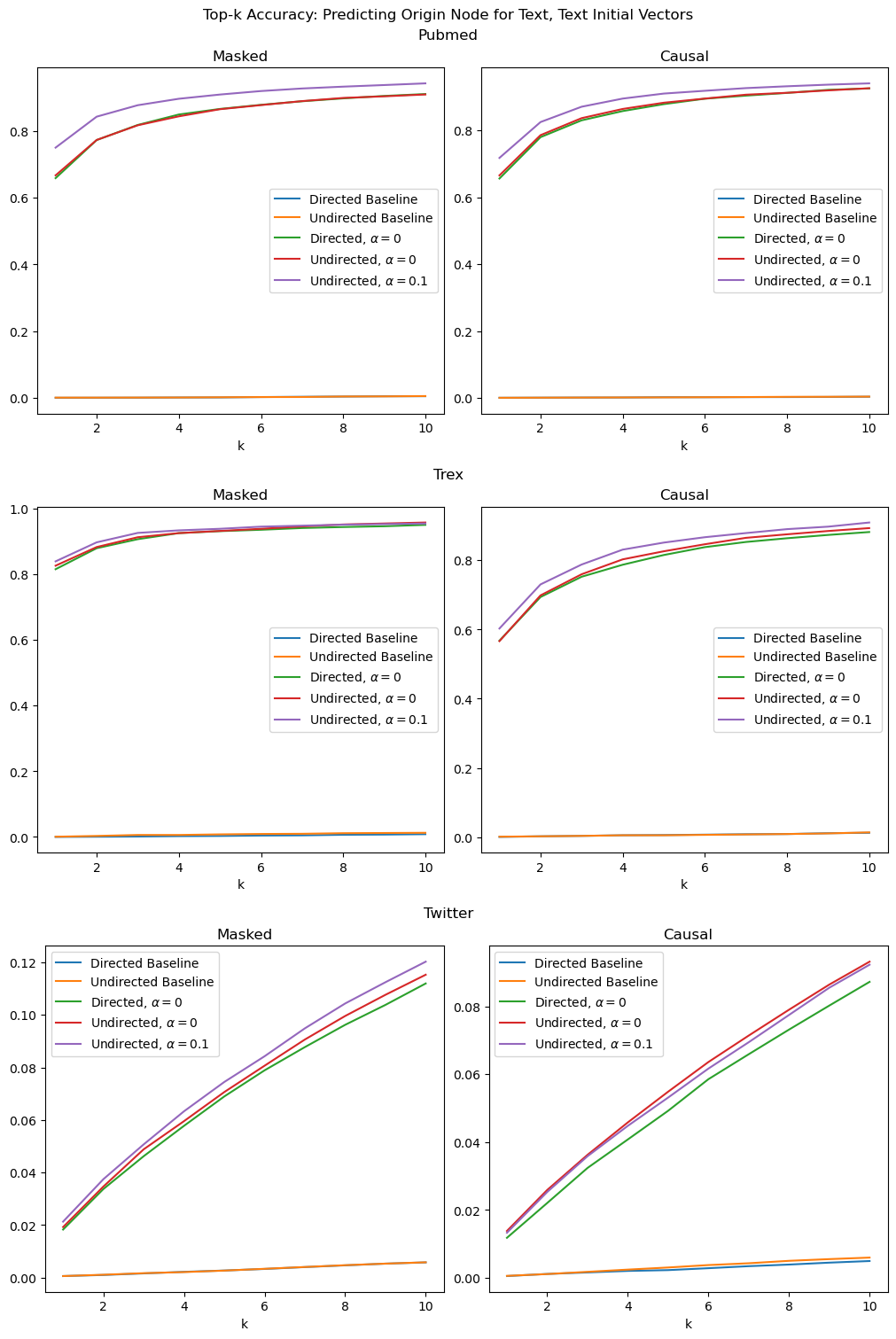}
    \caption{Top-\textit{k} accuracy on selection of the node which produced a text, for various values of $k$, as discussed in \autoref{subsec:appendix-embedding-geometry-retrieval}. ``Baseline'' indicates the use of separately pretrained embeddings, and other results are for models with various combinations of edge-direction use and graph-similarity information.}
    \label{fig:topk-acc-txt}
\end{figure*}

Results are shown in \autoref{fig:topk-acc-txt}. Top-$k$ accuracy is substantially higher than the separately-trained baseline for all models at all values of $k$. All differences are significant at the $p = 10^{-6}$ level according to a bootstrap test. Moreover, the top-k accuracies achieved are often high relative to the size of the datasets. With 1,996 articles (i.e., nodes) in the Pubmed test set, the best-performing model includes the correct article for a text snippet in its top 10 most similar articles (0.5\% of the test set) 94.3\% of the time.

\begin{table*}[!htb]
\centering
\setlength{\tabcolsep}{5pt}
\begin{tabular}{llcccccccccccc}
\toprule
 &  & \multicolumn{2}{c}{Age} & \multicolumn{2}{c}{Gender} & \multicolumn{2}{c}{Occupation} & \multicolumn{2}{c}{Region} \\
\cmidrule(lr){3-4} \cmidrule(lr){5-6} \cmidrule(lr){7-8} \cmidrule(lr){9-10}
&  & C & M & C & M & C & M & C & M \\
\midrule
\multirow[c]{3}{*}{Graph} & ConGraT ($\alpha = 0$) & 0.769 & 0.782 & 0.803 & \bfseries 0.832 & 0.993 & \bfseries 0.994 & 0.765 & 0.769 \\
 & ConGraT ($\alpha = 0.1$) & \bfseries 0.776 & \bfseries 0.801 & \bfseries 0.811 & 0.826 & \bfseries 0.993 & 0.993 & \bfseries 0.773 & \bfseries 0.774 \\
 \cmidrule(lr){2-10}
 & GAT & 0.767 & 0.767 & 0.791 & 0.791 & 0.991 & 0.991 & 0.706 & 0.706 \\
\midrule
\multirow[c]{5}{*}{Text} & ConGraT ($\alpha = 0$) & \bfseries 0.620 & 0.631 & 0.658 & \bfseries 0.667 & \bfseries 0.961 & \bfseries 0.962 & \bfseries 0.692 & \bfseries 0.689 \\
 & ConGraT ($\alpha = 0.1$) & 0.619 & \bfseries 0.635 & \bfseries 0.661 & 0.666 & 0.959 & 0.960 & 0.684 & 0.684 \\
 \cmidrule(lr){2-10}
 & LinkBERT & -- & 0.617 & -- & 0.661 & -- & 0.954 & -- & 0.606 \\
 & Social-LM & 0.566 & 0.567 & 0.602 & 0.608 & 0.921 & 0.909 & 0.582 & 0.572 \\
 & Unimodal LM & 0.610 & 0.613 & 0.649 & 0.655 & 0.948 & 0.945 & 0.587 & 0.598 \\
\bottomrule
\end{tabular}
\caption{Node classification performance on user traits from the Twitter dataset. (C = causal, M = masked.) Values are test-set AUCs. \textbf{Bold} values denote the best models in each experiment.
All differences between the best ConGraT model and the closest baseline are statistically significant ($p < 10^{-4}$) by a bootstrap test. These results are from models which use initial node representations (where applicable) based on the truncated SVD of the graph adjacency matrix rather than sentence embeddings.}
\label{tab:classification-svd-twitter}
\end{table*}

\section{Robustness Check: SVD-Based Initial Vectors}
\label{sec:appendix-svd}
In this section, we replicate the analysis described in \autoref{sec:experiments}, with a twist: instead of the sentence embeddings used there as initial node representations for those models which rely on them, we use vectors from the truncated SVD of the graph adjacency matrix. We train and evaluate entirely new models, in which all other properties of training, inference and datasets besides the choice of initial node vectors are the same as in the main text. Doing so provides an additional check on the soundness of our approach and gives evidence of its adaptability to various kinds of attributed graphs.

We truncate the SVD embeddings produced by scikit-learn's implementation \citep{pedregosaScikitlearnMachineLearning2011} to 768 dimensions, so that the embedding dimensionality is the same as for the sentence embeddings. To avoid leakage between train and test, we use the $V^*$ matrix from the train set to generate test-set embeddings (thus relying only on test-set nodes' connections to nodes in the training set).

\subsection{Node Classification}
\label{subsec:appendix-svd-classification}
Node classification performance is similar to that reported in the main text, with the best-performing ConGraT model outperforming all of our baseline models in all tested cases.

\paragraph{Twitter.}
\autoref{tab:classification-svd-twitter} shows AUCs for the demographic classification tasks. Results are similar to what we observe with sentence embeddings as initial node representations: the best-performing ConGraT model beats all baselines in evaluation with both graph and text embeddings.

Unlike in the maint text, however, we do not present results for the political-party outcome variable. This is because the SVD-based embeddings are too predictive: all graph models (and thus also all joint models) are able to perfectly separate the two classes. This phenomenon is a good example of the Twitter follow graph's powerful organizing principle of homophily: Users tend to be connected to other users who are similar to them \citep{barberaBirdsSameFeather2015}, in this case politically.

\paragraph{Pubmed and T-REx.}
AUCs for article category classification are shown in \autoref{tab:classification-svd-pubmed-trex}. As with the models using sentence embeddings, the best ConGraT model outperforms our baselines in all experiments, using both graph and text embeddings.

\begin{table}[htb]
\centering
\setlength{\tabcolsep}{5pt}
\footnotesize\begin{tabular}{llrrrr}
\toprule
 &  & \multicolumn{2}{c}{Pubmed} & \multicolumn{2}{c}{T-Rex} \\
\cmidrule(lr){3-4} \cmidrule(lr){5-6}
 &  & C & M & C & M \\
\midrule
\multirow[c]{3}{*}{\rotatebox[origin=c]{90}{Graph}} & ConGraT ($\alpha = 0$) & \bfseries 0.877 & 0.870 & 0.835 & 0.799 \\
 & ConGraT ($\alpha = 0.1$) & 0.868 & \bfseries 0.878 & \bfseries 0.854 & \bfseries 0.833 \\
\cmidrule(lr){2-6} 
 & GAT & 0.600 & 0.600 & 0.720 & 0.720 \\
\midrule
\multirow[c]{5}{*}{\rotatebox[origin=c]{90}{Text}} & ConGraT ($\alpha = 0$) & 0.955 & 0.954 & \bfseries 0.926 & 0.902 \\
 & ConGraT ($\alpha = 0.1$) & \bfseries 0.961 & \bfseries 0.955 & 0.923 & \bfseries 0.910 \\
\cmidrule(lr){2-6} 
 & LinkBERT & -- & 0.954 & -- & 0.906 \\
 & Social-LM & 0.858 & 0.878 & 0.890 & 0.851 \\
 & Unimodal LM & 0.931 & 0.943 & 0.908 & 0.892 \\
\bottomrule
\end{tabular}
\caption{Node classification performance on article labels of the Pubmed and T-REx datasets. Values are test-set AUCs. (C = causal, M = masked.) For T-REx, we show the average AUC over the category labels because the dataset is multilabel rather than multiclass. \textbf{Bold} values denote the best model in each experiment. All differences between the best ConGraT model and the closest baseline are statistically significant ($p < 10^{-4}$) by a bootstrap test. These results are from models which use initial node representations (where applicable) based on the truncated SVD of the graph adjacency matrix rather than sentence embeddings.}
\label{tab:classification-svd-pubmed-trex}
\end{table}

\subsection{Link Prediction}
\label{subsec:appendix-svd-link-prediction}
Link prediction results, given in \autoref{tab:link-prediction-svd}, are broadly similar to those in the main text and its \autoref{tab:link-prediction}. On all three datasets, we see ConGraT node encoders deliver much better than chance performance as zero-shot link predictors.

Performance is often very similar to the levels achieved with sentence embeddings, as on the Pubmed and undirected T-REx datasets. The most notable difference in relative performance vs baseline is on the Twitter dataset, where in fact ConGraT performance is quite close to that in \autoref{tab:link-prediction} --- the GAT baseline, however, performs much better with SVD-based embeddings.

\begin{table}[ht]
\centering
\setlength{\tabcolsep}{3.5pt}
\footnotesize\begin{tabular}{lcccccc}
\toprule
 &  &  & Pubmed & TRex & Twitter \\
 &  &  &  &  &  \\
\midrule
\multirow[c]{3}{*}{Masked} & \multirow[c]{2}{*}{$\alpha = 0$} & Undir. & 0.985 & 0.816 & 0.805 \\
 &  & Dir. & 0.955 & 0.663 & 0.826 \\
 & $\alpha = 0.1$ & Undir. & \bfseries 0.990 & \bfseries 0.882 & 0.790 \\
\midrule
\multirow[c]{3}{*}{Causal} & \multirow[c]{2}{*}{$\alpha = 0$} & Undir. & 0.976 & 0.886 & 0.793 \\
 &  & Dir. & 0.952 & 0.758 & 0.819 \\
 & $\alpha = 0.1$ & Undir. & \bfseries 0.984 & \bfseries 0.925 & 0.785 \\
\midrule
\multirow[c]{2}{*}{Baseline} & \multirow[c]{2}{*}{GAT} & Undir. & 0.866 & 0.839 & \bfseries 0.875 \\
 &  & Dir. & 0.864 & 0.719 & 0.838 \\
\bottomrule
\end{tabular}
\caption{Link prediction performance on each dataset. Values are test-set AUCs. \textbf{Bold} values denote the best model in each experiment. These results are from models which use initial node representations (where applicable) based on the truncated SVD of the graph adjacency matrix rather than sentence embeddings.}
\label{tab:link-prediction-svd}
\end{table}

\subsection{Language Modeling}
\label{subsec:appendix-svd-language-modeling}
Results here are similar to those presented in the main text, though ConGraT model performance declines slightly on the Pubmed dataset.

\begin{table}[!htb]
\centering
\setlength{\tabcolsep}{3.5pt}
\footnotesize\begin{tabular}{lccc}
    \toprule
    & Pubmed & T-REx & Twitter \\
    \midrule
    ConGraT (Causal, $\alpha = 0$) & 7.03 & 17.42 & 16.05 \\
    ConGraT (Causal, $\alpha = 0.1$) & 7.03 & \bfseries 15.62 & \bfseries 16.05 \\ \midrule
    Unimodal LM (Baseline) & \bfseries 6.98 & 16.84 & 16.44 \\
    \bottomrule
\end{tabular}
\caption{Language modeling performance of the ConGraT models with causal text encoders, vs. a unimodal LM baseline. Values are mean perplexity  (lower is better). \textbf{Bold} values are the best models on each dataset. All differences from the unimodal baseline are significant by a paired $t$-test at the $5\sigma$ level ($p < 5.7 \times 10^{-7}$). These results are from models which use initial node representations (where applicable) based on the truncated SVD of the graph adjacency matrix rather than sentence embeddings.}
\label{tab:lm-perplexity-svd}
\end{table}
\section{Additional Related Work}
\label{sec:appendix-related}
Many traditional NLP tasks focus on learning graph structures that exist within a text, such as dependency parsing \citep{kublerDependencyParsing2009} or grammar induction \citep{kleinGenerativeConstituentcontextModel2001, kimCompoundProbabilisticContextFree2019}. More recent lines of work have extended this focus on graph structures to knowledge graphs, where the graph structure is over knowledge-base entities which may appear in the text, and citation or social networks, where the relevant graph is the one between the entities which write or contain the texts. Social science applications have also frequently motivated approaches to learning from joint graph/text data.

\paragraph{Knowledge graphs.}
Work on knowledge graphs (KGs) uses graph representations to encode facts about the world \citep{hoganKnowledgeGraphs2022}, with real-world entities as nodes and edges used to encode inter-entity relationships. Previous works have presented models for jointly representing texts and KG entities, or entire graphs associated with each text \citep{toutanovaRepresentingTextJoint2015}, often focusing on question answering or reasoning \citep{zhangGreaseLMGraphREASoning2022, sunCoLAKEContextualizedLanguage2020, yasunagaQAGNNReasoningLanguage2021} or text generation \citep{keJointGTGraphTextJoint2021}. Our work differs in the architecture employed (a contrastive text/graph matching objective) and in the setting it is specialized for: text-attributed graphs, where the graph is over entities which produce (e.g., Twitter users) or contain (e.g., academic articles) their associated texts.

\paragraph{Applications.}
Many studies in the joint graph/text domain have focused on social-science questions in addition to machine-learning ones. Learning the community structure of social networks is a common application \citep{martinCommunity2vecVectorRepresentations2017, wallerQuantifyingSocialOrganization2021}, as is detecting hate speech, misinformation or fake news \citep{vijayaraghavanInterpretableMultiModalHate2019, chandraGraphbasedModelingOnline2020}. Some work has also focused on understanding political polarization or ideological differences between groups \citep{milbauerAligningMultidimensionalWorldviews2021, lyuUnderstandingPoliticalPolarization2022}.
\section{Dataset Details}
\label{sec:appendix-datasets}

\paragraph{Twitter.}
We created a Twitter dataset of 167,558 tweets posted by a set of 8,721 influential users in media, politics, and entertainment, and the follow graph among these users, which consists of 2,373,956 edges. We included up to 20 tweets per user in the dataset, sampled from each user's most recent 3,200 tweets as of May 9, 2021. We also collected certain demographic data about these users (region of residence, age, gender, politician vs. entertainer occupation, and political party) by matching them to Wikipedia and Ballotpedia\footnote{\url{https://ballotpedia.org/}}.

To obtain the age and gender of Twitter users, we connected the accounts to their corresponding Wikipedia pages and used Wikidata to infer those two features. Users also self-report locations in their Twitter bios; from these locations, we created four regional categories. Finally, we used data from Ballotpedia to label whether a user is a politician or not and to identify their political party. Note that politician status and party are derived in different ways, from different data fields, with politician status being defined more strictly. These variables, used as targets in node classification tasks, are broken down in \autoref{tab:dvs}.

\paragraph{Pubmed.}
We built from scratch a version of the popular Pubmed graph learning benchmark \citep{senCollectiveClassificationNetwork2008} that includes the titles and abstracts of each article; widely available versions of the dataset do not include any text. We began with the standard list of PMIDs for the articles in the dataset and fetched the title, abstract, and list of references using the Pubmed API. We kept directed citation edges only to other articles in the dataset. One PMID was not found in the Pubmed database and was thus left out. The final dataset includes 19,716 nodes, 61,110 edges, and 59,381 texts, including both titles and abstracts. The included articles are about diabetes and the standard node categories are from the Pubmed database: type-1 diabetes, type-2 diabetes, or experimental evidence.

\paragraph{T-REx.}
We used the articles in the T-REx corpus \citep{elsaharTRExLargeScale2018} of Wikipedia articles that were labeled with the ``Robots'' category or any of its descendant categories. From these categories, we constructed several binary target label sets for the T-REx prediction task. However, since the most commonly occurring category was only associated with 526 (roughly 5.7\%) of the articles, we expanded each article's labels to include both first and second level ancestors in the category hierarchy to obtain better class label balance. From the initial set of 1,433 unique categories, this expansion yielded a total of 6,643 unique categories, with the most frequent (``Spacecraft'') occurring on 1,342 articles. We then selected five categories to use as labels for separate binary prediction tasks, choosing frequent categories that generally had small overlap with each other (i.e. were associated with mostly disjoint document sets.) Note that not every data point in the dataset, then, received a label. The resultant categories we selected are listed in \autoref{tab:dvs}.

\begin{table}[!htb]
\centering
\footnotesize\begin{tabular}{@{}llll@{}}
\toprule
\textbf{Dataset}          & \textbf{Feature}            & \textbf{Category} & \textbf{\# Nodes} \\
\midrule
\multirow{13}{*}{Twitter}   & \multirow{4}{*}{Region}           & Midwest           & 64                \\
                            &                                   & Northeast         & 1207              \\
                            &                                   & South             & 1411              \\
                            &                                   & West              & 1100              \\
\cmidrule(lr){2-4}
                            & \multirow{3}{*}{Age}              & 19-39             & 575               \\
                            &                                   & 40-49             & 2216              \\
                            &                                   & \textgreater{}=65 & 844               \\
\cmidrule(lr){2-4}
                            & \multirow{2}{*}{Gender}           & Female            & 1586              \\
                            &                                   & Male              & 2495              \\
\cmidrule(lr){2-4}
                            & \multirow{2}{*}{Occupation}       & Non-politician    & 8271              \\
                            &                                   & Politician        & 434               \\
\cmidrule(lr){2-4}
                            & \multirow{2}{*}{Party}            & Democrat          & 241               \\
                            &                                   & Republican        & 193               \\
\midrule
\multirow{3}{*}{Pubmed}     & \multirow{3}{*}{Article Type}     & Experimental      & 4103              \\
                            &                                   & Type I            & 7875              \\
                            &                                   & Type II           & 7738              \\
\midrule
\multirow{5}{*}{T-REx}      & \multirow{5}{*}{Wiki Category}    & Robots            & 666               \\
                            &                                   & Rockets           & 843               \\
                            &                                   & Sci-Fi            & 712               \\
                            &                                   & Spacecraft        & 1342              \\
                            &                                   & Space Telescopes  & 701               \\

\bottomrule
\end{tabular}
\caption{Breakdowns of the dependent variables for node classification experiments on the three datasets.}
\label{tab:dvs}
\end{table}

\paragraph{Splitting.}
We divide the datasets into a 70\% train set, 10\% validation set, and 20\% test set, splitting at the node level so that every text associated with a given node is in the same split. Because evaluation is inductive, any graph edges which cross split boundaries are dropped.
\section{Model Architectures and Training Details}
\label{sec:appendix-models-training}
We estimate that training all of our joint and baseline models together used 263 hours of GPU time. Because the assumptions made for this value are conservative, the actual value is likely slightly less.

\subsection{ConGraT Models}
\label{subsec:appendix-training-congrat}
We trained all ConGraT models on either a single NVIDIA RTX A6000 GPU or a single NVIDIA A100 GPU. For masked LM experiments, we used the pretrained \texttt{all-mpnet-base-v2} model \citep{songMPNetMaskedPermuted2020} from the sentence-transformers toolkit \citep{reimersSentenceBERTSentenceEmbeddings2019}, which has 12 layers of 12 heads each, producing 768-dimensional embeddings. It was pretrained constrastively on several corpora from similar domains to those we consider here,\footnote{See the model card for details: \url{https://huggingface.co/sentence-transformers/all-mpnet-base-v2}.} making it a good match for our work. Our causal LM experiments used the pretrained \texttt{distilgpt2} model \citep{sanhDistilBERTDistilledVersion2019}, distilled from the larger GPT-2 model \citep{radfordLanguageModelsAre2019}, with 6 layers of 12 heads each, producing 768-dimensional embeddings.\footnote{Again see the model card for more details: \url{https://huggingface.co/distilgpt2}.} For the graph node encoder, all models used a graph attention network (GAT) \citep{velickovicGraphAttentionNetworks2018} with 3 layers and 2 attention heads in each layer. As in a standard transformer, each graph convolutional layer is separated from the next by a linear layer, with layer normalization \citep{baLayerNormalization2016} applied afterwards. Hidden representations are 64-dimensional, and the final output vectors are 768-dimensional so that baseline model outputs have the same shape as language model outputs.

Parameter counts are as follows: \texttt{distilgpt2}, 81.9 million; \texttt{all-mpnet-base-v2}, 109.4 million; our GAT encoder, 199.7 thousand. The jointly trained models, including the adapter layers after the text and graph encoders, have 83.9 million parameters (causal / \texttt{distilgpt2}) and 110.9 million parameters (masked / \texttt{all-mpnet-base-v2}).

Training is sensitive to the learning rate; we found that a good compromise between speed of training and instability was a value of \texttt{1e-4}. At a variety of learning rates, there were also intermittent large spikes in the norm of the gradient, which derailed training unless the gradients were clipped. We clipped the gradient at each step to a norm of 1. In order to reduce memory consumption and fit larger batches onto a single GPU, we used 16-bit mixed precision training \citep{micikeviciusMixedPrecisionTraining2018}. We encountered numerical overflow problems with FP16, however, related to large gradient values at certain layers, and found it necessary to reduce the init-scale parameter of the gradient scaler from its default value of $2^{16}$ to 256 in order to avoid overflow. We initialized the log-temperature parameter $\tau$ to $3.5$ and constrained it to be within $(-\log 100, +\log 100)$ in order to avoid training instability. We trained all models with PyTorch \citep{paszkePyTorchImperativeStyle2019} and pytorch-lightning \citep{falconPyTorchLightningPytorchlightningRelease2020}, also using pytorch-geometric \citep{feyFastGraphRepresentation2019} for graph encoders and GAT baselines, and Huggingface Transformers \citep{raffelExploringLimitsTransfer2020} for textual baselines and text encoders.

We also found that performance suffers if each batch is not unique on nodes (i.e., if each node has multiple texts, only one text per node can be in any given batch). We experimented with simply dropping duplicates from uniformly sampled batches, but this discarded too much data. Instead, we randomly sorted the texts on each epoch so as to minimize the number of within-batch duplicates (assuming minibatches are taken consecutively from the sorted dataset), and dropped any remaining duplicates.

Finally, because the objective is batch-wise contrastive, the problem becomes quadratically more difficult as the batch size increases. We used the largest batch size we could consistently fit into available hardware, but future work should explore the question of returns to scale.

All models used the AdamW optimizer \citep{loshchilovDecoupledWeightDecay2019} with $\beta$ values of (0.9, 0.999) and without weight decay. All joint models used a probability of 0.3 for dropout applied to text and node embeddings. Learning rates and batch sizes for our various models are shown in \autoref{tab:batch-size-lr}.

\begin{table*}[!htb]
\centering
\begin{tabular}{c|c|c|c}
    \toprule
     Model or Model Family & Batch Size & Base LR & LR Schedule \\
     \midrule
     ConGraT & 36 & \texttt{1.0e-4} & Constant LR \\
     LM Baseline & 36 & \texttt{5.0e-5} & Linear 10\% warmup \\
     SocialLM & 36 & \texttt{5.0e-5} & Linear 10\% warmup \\
     LinkBERT & 36 & \texttt{5.0e-5} & Linear 10\% warmup \\
     \midrule
     GNN AE (Baseline), Twitter, Dir. & n/a & \texttt{1.0e-2} & Constant LR \\
     GNN AE (Baseline), Twitter, Undir. & n/a & \texttt{1.0e-2} & Constant LR \\
     GNN AE (Baseline), T-REx, Dir. & n/a & \texttt{1.0e-2} & Constant LR \\
     GNN AE (Baseline), T-REx, Undir. & n/a & \texttt{1.0e-2} & Constant LR \\
     GNN AE (Baseline), Pubmed, Dir. & n/a & \texttt{1.0e-2} & Constant LR \\
     GNN AE (Baseline), Pubmed, Undir. & n/a & \texttt{1.0e-2} & Constant LR \\
     \midrule
     GNN AE (SVD), Twitter, Dir. & n/a & \texttt{1.0e-2} & Constant LR \\
     GNN AE (SVD), Twitter, Undir. & n/a & \texttt{1.0e-2} & Constant LR \\
     GNN AE (SVD), T-REx, Dir. & n/a & \texttt{1.0e-3} & Constant LR \\
     GNN AE (SVD), T-REx, Undir. & n/a & \texttt{1.0e-3} & Constant LR \\
     GNN AE (SVD), Pubmed, Dir. & n/a & \texttt{1.0e-3} & Constant LR \\
     GNN AE (SVD), Pubmed, Undir. & n/a & \texttt{1.0e-3} & Constant LR \\
     \bottomrule
\end{tabular}
\caption{Batch sizes and learning rates for all models.  (AE = autoencoder.) Except for the GNN baseline's learning rate, where we tried both \texttt{1.0e-2} and \texttt{1.0e-3} and found large dataset-specific effects on performance, all models listed in the same model family use the same parameter settings for all datasets. In particular, all ConGraT models, whether directed or undirected, with $\alpha = 0$ or $\alpha = 0.1$, and causal or masked encoders, used the same batch size and learning rate. GNN baselines do not list a batch size because the entire graph is processed at once.}
\label{tab:batch-size-lr}
\end{table*}

\subsection{Unimodal Baselines}
\label{subsec:appendix-training-unimodal}
To better understand the effects of multi-modal pretraining, we also trained unimodal models, either language models or graph attention transformers, and evaluated these unimodal models on the downstream tasks. For textual models, we fine-tuned pretrained \texttt{all-mpnet-base-v2} and \texttt{distilgpt2} on the training splits of the evaluation datasets. Language models were fine-tuned for 3 epochs. For graph models, we trained graph attention network (GAT) models to do non-variational graph autoencoding \citep{kipfVariationalGraphAutoEncoders2016}, also known as link prediction, on the network structure of the evaluation datasets. GAT models were trained from between 30 to 100 epochs with early stopping based on validation AUC, with patience of 3 epochs and minimum delta of 0.01. We compare these unimodal baselines against ConGraT. Parameter counts for the text and graph baselines are the same as reported for the appropriate modality's joint encoder in \autoref{subsec:appendix-training-congrat}. Batch sizes and learning rates, as for joint models, are reported in \autoref{tab:batch-size-lr}. Our unimodal baselines were trained on NVIDIA RTX A6000 GPUs, or on up to four NVIDIA GTX 1080 Ti GPUs.

\subsection{Social-LM}
\label{subsec:appendix-training-socialbert}
We implemented a baseline Social-LM, as a modified version of SocialBERT\footnote{The SocialBERT authors did not publish their code.} \citep{karpovSocialBERTTransformersOnline2022a} (also very closely related to LMSOC \citep{kulkarniLMSOCApproachSocially2021}), which uses pretrained, frozen node embeddings to prime language model pretraining. Specifically, we added a special node token [G] at the beginning of texts and used the pretrained GAT model to obtain the corresponding node embedding paired with each tweet or article, which was used to replace the token embedding for [G]. During the language model pretraining, we froze the node embeddings and only fine-tuned the language model to generate texts conditioned on the node embeddings. Our Social-LM implementation has some key differences from SocialBERT and LMSOC: (1) for masked LM experiments, we used \texttt{all-mpnet-base-v2} to replace BERT, to be consistent with other experiments for a fair comparison; (2) we also experimented with a causal language model \texttt{distilgpt2} under the Social-LM baseline, whereas LMSOC and SocialBERT only used the masked language model BERT; (3) we injected the node embedding as the zero token embedding of texts as SocialBERT suggests, whereas LMSOC appends the node embedding at the end. We adopted the zero token injection approach because the same strategy is adaptable for both causal and masked language modeling, while last token injection does not work for causal LMs like \texttt{distilgpt2}; (4) we used our unimodal GAT model trained on the graph autoencoding task to generate node embeddings for each tweet or article, whereas LMSOC uses node2vec and SocialBERT uses vectors from SVD and Deep Walk. We used the GAT in order to be consistent with ConGraT and the unimodal baseline, to ensure that the comparisons were fair, and because it was likely to be a stronger baseline than using SVD. Social-LM models were fine-tuned for 3 epochs with the same hyperparameters used for the language modeling baseline, and have the same number of parameters as \texttt{all-mpnet-base-v2}, our masked LM baselines and the joint masked text encoders.

\subsection{LinkBERT}
\label{subsec:appendix-training-linkbert}
We implemented and trained LinkBERT \citep{yasunagaLinkBERTPretrainingLanguage2022} as described in the original paper, with the only difference being that we used the same  \texttt{all-mpnet-base-v2} architecture as the other baseline models (instead of BERT-Base) in order to maintain consistency across experiments. We initialized weights from the pretrained \texttt{all-mpnet-base-v2} model from sentence-transformers, and fine-tuned it on the masked language modeling (MLM) and document relation prediction (DRP) tasks for 3 epochs. Hyperparameters used for training are listed in \autoref{tab:batch-size-lr}. Note that because of its MLM training objective, we used LinkBERT as a baseline for masked language model variants of ConGraT only. All LinkBERT models have the same number of parameters as \texttt{all-mpnet-base-v2}, as the DRP head is dropped at inference time.

We created training instances for LinkBERT by sampling contiguous, linked, or random text segment pairs for the DRP training objective from each dataset, with the three options appearing uniformly (1/3, 1/3, 1/3). For the Pubmed and Twitter datasets, we sampled 100,000 text pairs for each category, for a total of 300,000 pairs. For T-REx, which is a substantially smaller dataset, we sampled 10,000 text pairs for each category, for a total of 30,000 pairs. Text pairs consisted of anchor text segment $X_A$ and paired text segment $X_B$: $(X_A, X_B)$. The specific methods we used to sample pairs for each dataset were as follows:

\paragraph{Pubmed.}
Text segments in each pair consisted of individual sentences from the abstracts of each article in the dataset. Anchor segments $X_A$ were taken by sampling a random abstract, then sampling a random sentence from that abstract. For contiguous pairs, $X_B$ was chosen as the sentence immediately following $X_A$ in the abstract ($X_A$ could not be the last sentence of the abstract). For linked pairs, $X_B$ was chosen as a random sentence from the abstract of one of the articles that was connected to $X_A$'s corresponding article in the citation graph. For random pairs, $X_B$ was chosen as a random sentence from an abstract whose article was not connected to $X_A$'s corresponding article in the citation graph.

\paragraph{T-REx.}
Text segments in each pair consisted of individual sentences from the introductory paragraphs of each article in the dataset. Anchor segments $X_A$ were taken by sampling a random article, then sampling a random sentence from that article's introductory paragraphs. For continuous pairs, $X_B$ was chosen as the sentence immediately following $X_A$, with the same restriction as in Pubmed that $X_A$ could not be the last sentence. For linked pairs, $X_B$ was chosen as a random sentence from the introductory paragraphs of one of the articles connected to $X_A$'s corresponding article in the link graph. For random pairs, $X_B$ was chosen as a random sentence from an article not connected to $X_A$'s corresponding article in the link graph.

\paragraph{Twitter.}
Twitter has a different graph-text structure than Pubmed and T-REx; rather than the nodes consisting of texts themselves, the nodes are users who can each produce multiple tweets. Therefore, the notion of what constitutes continuous or linked text segments (tweets) is less clearly defined. We defined these relationships as follows. For contiguous pairs, we sampled a random tweet as $X_A$, and chose $X_B$ as a different tweet from the same user as $X_A$. For linked pairs, we sampled $X_A$ from the set of tweets that \textit{mentioned} other users that were present in our dataset. Then, $X_B$ was chosen as a random tweet from the mentioned user. Random pairs were taken by randomly sampling two tweets from different users to use as $X_A$ and $X_B$.

\subsection{Node Classification Methodology}
\label{subsec:appendix-training-downstream}
We use the standard scikit-learn \citep{pedregosaScikitlearnMachineLearning2011} implementation of logistic/softmax regression with the default L2 regularization, balancing our sometimes very imbalanced classification problems by downsampling before fitting. For performance reasons we use the \texttt{liblinear} solver for problems with no more than 5000 training data points and the \texttt{saga} solver otherwise. To ensure convergence, we increase the maximum iterations for the solvers from the default of 100 to 10000.
\section{Licenses and Terms of Use}
\label{sec:terms-of-use}
All software and pretrained models we used were available under open-source licenses which permit our use for research purposes. Our non-Twitter datasets were available under Creative Commons or other licenses allowing research use. We have access to Twitter data pursuant to an agreement with Twitter permitting use of data for research and publication. The agreement permits releasing the tweet IDs, which can be used to get the corresponding tweets from the public Twitter API. Along with the tweet IDs, we plan to release the demographic data collected from Wikipedia and Ballotpedia. Our code and datasets, when released upon publication, will be subject to an open-source license allowing use for research purposes.

\end{document}